%% file: acl2023.tex
\pdfoutput=1
\documentclass[11pt]{article}

\usepackage{ACL2023}
\input{preamble.tex}
\usepackage{times}
\usepackage{latexsym}
\usepackage{booktabs}
\usepackage{multirow}
\usepackage{makecell}
\usepackage{xspace}

\usepackage[T1]{fontenc}
\usepackage[utf8]{inputenc}

\usepackage{microtype}

\usepackage{inconsolata}

\title{Models of reference production: How do they withstand the test of time?}

\author{Fahime Same\textsuperscript{$\heartsuit$}, 
Guanyi Chen\textsuperscript{$\spadesuit$}, \and
Kees van Deemter\textsuperscript{$\spadesuit$}\\
\textsuperscript{$\heartsuit$}Department of Linguistics, University of Cologne\\
\textsuperscript{$\spadesuit$}Department of Information and Computing Sciences, Utrecht University\\
\texttt{f.same@uni-koeln.de, g.chen@ccnu.edu.cn, c.j.vandeemter@uu.nl}}

\begin{document}
\maketitle
\begin{abstract}
In recent years, many NLP studies have focused solely on performance improvement. In this work, we focus on the linguistic and scientific aspects of NLP. We use the task of generating referring expressions in context (REG-in-context) as a case study and start our analysis from GREC, a comprehensive set of shared tasks in English that addressed this topic over a decade ago. We ask what the performance of models would be if we assessed them (1) on more realistic datasets, and (2) using more advanced methods. 
We test the models using different evaluation metrics and feature selection experiments. We conclude that GREC can no longer be regarded as offering a reliable assessment of
models' ability to mimic human reference production, because the results are highly impacted by the choice of corpus and evaluation metrics.
Our results also suggest that pre-trained language models are less dependent on the choice of corpus than classic Machine Learning models, and therefore make more robust class predictions.
\end{abstract}

\input{sec/intro.tex}

\input{sec/grec.tex}

\input{sec/rq.tex}
\input{sec/algorithm.tex}
\input{sec/dataset.tex}
\input{sec/evaluation.tex}
\input{sec/analysis.tex}

\input{sec/discussion.tex}

\paragraph*{\textit{Ethics Statement:}}

Regarding potential biases, in addition to the biases present in text-based datasets, biases can also be introduced by the pre-trained language models~\cite{bender2021dangers} used in this work. In other words, the REG algorithms we developed in this study may make different predictions with respect to different genders, for instance. In the future, we plan to investigate this phenomenon and find ways to mitigate it.

\paragraph*{\textit{Supplementary Materials Availability Statement:}}
All associated data, source code, output files, scripts, documentation, and other relevant material to this paper are publicly available and can be accessed on our GitHub repository: \url{https://github.com/fsame/REG_GREC-WSJ}, \href{https://doi.org/10.5281/zenodo.8182689}{DOI: 10.5281/zenodo.8182689}.

\paragraph*{\textit{Acknowledgements:}}
We thank the anonymous reviewers for their helpful comments.
Fahime Same is supported by the German Research Foundation (DFG)– Project-ID 281511265 – SFB 1252 ``Prominence in Language”.
\bibliography{anthology,custom}
\bibliographystyle{acl_natbib}

\newpage
\appendix
\input{sec/appendix}

\end{document}

%% file: preamble.tex
\usepackage{paralist}
\usepackage{tikz}
\usepackage{dblfloatfix}
\usepackage{xspace}

\usepackage{microtype}

\usepackage{amsmath}

\usepackage{lingmacros}

\newcommand{\bert}{\texttt{BERT}\xspace}
\newcommand{\roberta}{\texttt{RoBERTa}\xspace}
\newcommand{\udel}{\texttt{UDel}\xspace}
\newcommand{\cnts}{\texttt{CNTS}\xspace}
\newcommand{\isg}{\texttt{IS-G}\xspace}
\newcommand{\osu}{\texttt{OSU}\xspace}
\newcommand{\icsi}{\texttt{ICSI}\xspace}
\newcommand{\msr}{\textsc{msr}\xspace}
\newcommand{\negc}{\textsc{neg}\xspace}
\newcommand{\wsj}{\textsc{wsj}\xspace}

%% file: sec/intro.tex
\section{Introduction}

NLP research can have different aims. Some NLP research focuses on developing new algorithms or building practical NLP applications. Another line of NLP work constructs computational models that aim to explain human language and language use; this line of work has been dubbed \textit{NLP-as-Science}~\citep{van2022role}. Among other things, NLP-as-Science demands that we ask ourselves to what extent NLP research findings generalise along a range of dimensions.

In addition to the practical applications of Referring Expression Generation~\citep[REG, ][]{reiter-2017-commercial}, REG is also one of the typical tasks in NLP-as-Science, where REG algorithms are built to model and explain the reference production of human beings \cite{krahmer-van-deemter-2012-computational, van2016computational}. In the computational linguistics and cognitive science community, REG can be divided into two distinct tasks: \emph{one-shot REG}, finding a referring expression (RE) to single out a referent from a set, and \emph{REG-in-context}, generating an RE to refer to a referent at a given point in a discourse. 

In a classic setup, REG-in-context is often approached in two steps: The first is to decide on the form of an RE at a given point in the discourse, and the second is to decide on its content. Many researchers have been interested in the first sub-task, referential form selection: the task to decide which referential form (e.g., pronoun, proper name, description, etc.) an RE takes~\citep{mccoy1999generating, henschel2000pronominalization, kibrik2016referential}. Nearly 15 years ago, \citet{belz-etal-2008-grec} introduced the GREC shared tasks and a number of English REG corpora with two goals: (1) assessing the performance of computational models of reference production~\citep{belz2009generating}, and (2) understanding the contribution of linguistically-inspired factors to the choice of referential form~\citep{greenbacker2009feature, kibrik2016referential, same-van-deemter-2020-linguistic}.

15 years have passed since the GREC challenge was organised, and many new models and corpora have been proposed in the meantime (e.g., \citet{castro-ferreira-etal-2018-neuralreg,cunha-etal-2020-referring}, and \citet{same-etal-2022-non}). We, therefore, decided that it was time to ask, in the spirit of NLP-as-Science, how well the lessons that GREC once taught our research community %about human RE use 
hold up when scrutinised in light of all these developments. In other words, we will investigate to what extent the findings from GREC can be {\em generalised} to other corpora and other models. 

To this end, we pursue the following objectives: (1) We extend GREC by testing its REG algorithms not only on the GREC corpora but also on a corpus that was not originally considered and that has a different genre, namely the Wall Street Journal (WSJ) portion of OntoNotes~\citep{hovy2006ontonotes, weischedel2013ontonotes}; (2) We fine-tune pre-trained language models on the task of REG-in-context and assess them in the GREC framework. 

In Section \ref{sec:grec}, we detail the GREC shared tasks and introduce the corpora used in GREC. 
Section~\ref{sec:rq} spells out our research questions.
In Section~\ref{sec:algorithm} and Section~\ref{sec:corpus}, we introduce the algorithms and corpora that we use. Section~\ref{sec:evaluation} reports the performance of each algorithm on each corpus, followed by analyses in Section~\ref{sec:analysis}.
Section~\ref{sec:discussion} will discuss our findings and draw some lessons.

%% file: sec/grec.tex
\section{The GREC Shared Tasks}\label{sec:grec}

In this section, we summarise the GREC task, the corpora used by GREC, and its conclusions.

\subsection{The GREC Task and its Corpora} \label{subsec:greccorpora}

According to \citeauthor{belz2009generating}, ``\emph{the GREC tasks are about how to generate appropriate references to an entity in the context of a piece of discourse longer than a sentence}" \citeyearpar[p.~297]{belz2009generating}. The main task 
was to predict the referential form, namely whether to use a pronoun, proper name, description or an empty reference at a given point in discourse.

The GREC challenges use two corpora, both created from the introductory sections of Wikipedia articles: (1) GREC-2.0 (henceforth \msr, as it was used in the GREC-MSR shared tasks of 2008 and 2009) consists of 1941 introductory sections of the articles across five domains (people, river, mountain, city, and country); and (2) GREC-People (henceforth \negc as it was used in the GREC-NEG shared task in 2009) contains 1000 introductory sections from Wikipedia articles about composers, chefs, and inventors. Here is an example from \negc:

\enumsentence{\underline{\textbf{David Chang}} (born 1977) is a noted American chef. \underline{\textbf{He}} is chef/owner of Momofuku Noodle Bar, Momofuku Ko and Momofuku Ssäm Bar in New York City. \underline{\textbf{Chang}} attended Trinity College, where \underline{\textbf{he}} majored in religious studies. In 2003, \underline{\textbf{Chang}} opened \underline{\textbf{his}} first restaurant, Momofuku Noodle Bar, in the East Village.}\label{ex:davidchang}

A key difference between \msr and \negc lies in their RE annotation practices. In \msr, only those REs that refer to the main topic of the article are annotated, while in \negc, mentions of all \textit{human} referents are annotated. For instance, in a document about David Chang, \msr will only annotate REs referring to David Chang, while \negc will include annotations for all human referents, including David Chang and others.

\subsection{REG Algorithms Submitted to GREC} \label{subsec:systems}

Various REG algorithms were submitted to the GREC challenges. 
These consist of feature-based ML algorithms: \cnts~\citep{hendrickx-etal-2008-cnts}, 
\icsi~\citep{favre-bohnet-2009-icsi}, \isg~\citep{bohnet-2008-g}, \osu~\citet{jamison-mehay-2008-osu} and \udel~\citet{greenbacker-mccoy-2009-udel}, and an algorithm that mixes feature-based ML and rules: \texttt{JUNLG}~\citep{gupta2009junlg}.
Table \ref{tab:grecsystems} presents the details of each model, including the ML method, and the original reported accuracy on \msr (cf. \citet{belz2009generating} for details).

\input{tab/grec_results.tex}

\subsection{Feature Selection}

The GREC Tasks were designed to find out \emph{what kind of information is useful for making choices between different kinds of referring expressions in context} \citep[p. 297]{belz2009generating}. However, the original paper does not consider the factors that contributed to the RE choice in the systems submitted to GREC. In a follow-up study, \citet{greenbacker2009feature} conducted a feature selection study informed by psycholinguistics. They experimented with various feature subsets derived from their system, known as UDel, which had previously been submitted to the GREC. Additionally, they incorporated selected features from another REG system, CNTS \citep{hendrickx-etal-2008-cnts}, into their study. They show that features motivated by psycholinguistic studies and certain sentence construction features have a positive impact on the performance of REG models. Follow-up feature-selection studies including \citet{kibrik2016referential} and \citet{same-van-deemter-2020-linguistic} also emphasise the contribution of factors such as recency and grammatical role to the choice of RE form.

%% file: tab/grec_results.tex
\begin{table}[tbp]
\centering
\begin{tabular}{lccc}
\hline
Name & GREC ST & ALG & Acc   \\ \hline
\udel & \msr’09  & C5.0   & 77.71 \\ 
\icsi & \msr’09  & CRF    & 75.16 \\ 
\cnts & \msr’08  & MBL    & 72.61 \\ 
\isg & \msr’08  & MLP    & 70.78 \\ 
\osu & \msr’08  & MaxEnt & 69.82 \\ 
\texttt{JUNLG} & \msr’09 & Rule & 75.40 \\
\hline
\end{tabular}
\caption{\label{tab:grecsystems}
An overview of the algorithms submitted to GREC. The first column contains the name of the respective algorithm. The column GREC ST presents the name of the \msr shared task to which the algorithm was submitted. The third column, ALG, lists the algorithms used, where abbreviations from top to bottom are C5.0 decision tree, conditional random field, memory-based learning, multi-layer perceptron, maximum entropy, and frequency-based rules. The fourth column, Acc, reports the original accuracy of the algorithms, as reported in \citet{belz2009generating}. Note that \udel, \icsi, and \texttt{JUNLG} were submitted to both the \msr’08 and \msr'09 shared tasks, and we only present the newest results here.
}
\end{table}

%% file: sec/rq.tex
\section{Research Questions} \label{sec:rq}

15 years after the GREC shared tasks, we were curious to know to what extent the conclusions from GREC still ``stand''. We, therefore, came up with the following research questions.

In the first place, we are interested in \emph{the impact of the choice of corpus on the performance of REG algorithms} ($\mathcal{R}_1$). GREC uses only the introductory part of Wikipedia articles (see Section~\ref{sec:grec}), which represents only one genre of human language use. Considering that a good REG algorithm needs to model the general use of reference, a better evaluation framework should include texts from multiple genres. Therefore, we also include the WSJ corpus in the study
(see Section~\ref{sec:corpus} for more details) and conduct a correlation analysis to quantify how the choice of corpus impacts the evaluation results. 

Second, previous studies suggested that classic machine learning (ML) based REG algorithms perform on par with most recent neural methods~\citep{same-etal-2022-non}. However, their study has three limitations: (1) they did not incorporate pre-trained language models (PLMs); (2) they focused on the surface forms of REs, which partly depend on the performance of surface realisation; (3) they did not assess the models based on the intuition that a model with good explanatory power should be less influenced by the choice of corpus. Therefore, we adopt PLMs to the task of REG-in-context (see Section~\ref{sec:algorithm} for more details) and investigate \emph{how good is the explanatory power of PLM-based REG models compared to classic ML-based models} ($\mathcal{R}_2$) using the enhanced GREC framework.

Finally, as previously mentioned, one of the primary theoretical objectives of GREC was to computationally explore the contribution of factors that originate from linguistic studies to the choice of referential forms. It is reasonable to expect that such contributions may change depending on the choice of corpus. In this study, we conduct an importance analysis to investigate \emph{whether the importance ranking of linguistic factors changes when we use different corpora} ($\mathcal{R}_3$).

%% file: sec/algorithm.tex
\section{REG Algorithms} \label{sec:algorithm}

In what follows, we introduce the REG algorithms that are considered in this study.

\subsection{ML-based REG} 

For this study, we have narrowed our focus to feature-based ML algorithms that predict the type of RE. Consequently, we reconstruct five ML-based REG algorithms, namely \udel, \icsi, \cnts, \isg, and \osu, along with their respective feature sets, while excluding \texttt{JUNLG}. Note that we implement \cnts slightly differently from \citet{hendrickx-etal-2008-cnts}. Concretely, \citet{hendrickx-etal-2008-cnts} have mentioned that they have used the TiMBL package \citep{daelemans2007timbl} for implementing the Memory Based Learning algorithm. Instead, we implemented the k-Nearest Neighbors algorithm. According to \citet{daelemans2007timbl}, Memory Based Learning is the direct descendant of k-Nearest Neighbors. More information on the implementation of these models can be found in Appendix \ref{sec:appendixML}.

\subsection{PLM-based REG}

\begin{figure}[t]
    \centering
    \includegraphics[scale=0.35]{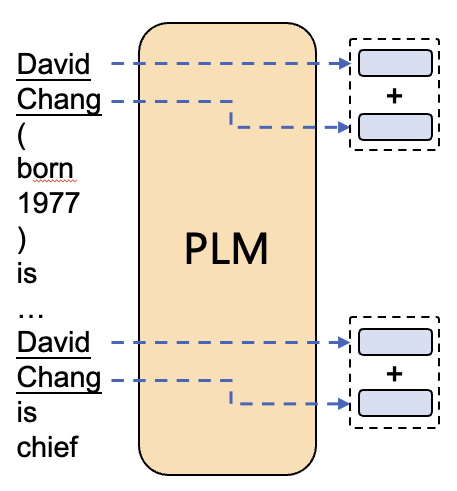}
    \caption{Illustration of the PLM-based REG Algorithm.}
    \label{fig:plm}
\end{figure}

Deep learning approaches have been used in many previous works on REG~\citep{castro-ferreira-etal-2019-neural, cao-cheung-2019-referring, cunha-etal-2020-referring, chen-etal-2021-neural-referential}. Different from previous work\footnote{Note that \citet{chen-etal-2021-neural-referential,chen2023neural} also leveraged a PLM, but did not fine-tune it. Instead, they used the word representations from the PLM as static inputs to an RNN and made predictions using the RNN.}, we fine-tune PLMs on REG corpora in this study. 

To fine-tune PLMs on REG corpora, we began by pre-processing each corpus using the same paradigm as described by ~\citet{cunha-etal-2020-referring}. More precisely, each referent in a given document was replaced with its corresponding proper name. For example, all underlined REs in Example (1) were replaced by ``David Chang''. Subsequently, as depicted in Figure~\ref{fig:plm}, we fed the data into a PLM, and, for each referent (e.g., ``David Chang'' ), we extracted the representations of its first token and its last token and summed them. The final representations were then sent to a fully connected layer for predicting the RE forms. In this study, we use \texttt{BERT} and \texttt{RoBERTa} (see section \ref{sec:implementation} for more details).

%% file: sec/dataset.tex
\section{REG Corpora} \label{sec:corpus}

In the following, we explain the corpora used in this work. These corpora are English-language corpora.
\subsection{The \msr and \negc Corpora}
In the current study, we only use the articles from the training sets of these corpora (see the number of documents in Table \ref{tab:corpora}). Following the same approach as \citet{castro-ferreira-etal-2018-neuralreg}, we created a version of the GREC corpora for the End-to-end (E2E) REG modelling. For the classic ML models, we reproduced the models using the feature sets from the studies mentioned in Section \ref{subsec:systems}.

\subsection{The \wsj Corpus}

As mentioned earlier, the WSJ portion of the OntoNotes corpus \citep{weischedel2013ontonotes} is our third data source.\footnote{We used Ontonotes 5.0 licensed by the Linguistic Data Consortium (LDC) \url{https://catalog.ldc.upenn.edu/LDC2013T19}.} We use the version of the corpus that \citet{same-etal-2022-non} developed for E2E REG modeling.\footnote{Note that \wsj was used in~\citet{same-etal-2022-non}, but no corpus analysis or comparison was provided.} Since empty pronouns are not annotated in \wsj, we decided to also exclude them from the two GREC corpora and focus on a 3-label classification task. The labels considered in this study are \emph{pronoun}, \emph{description},  and \emph{proper name}. Table \ref{tab:corpora} presents a detailed overview of these corpora.

\begin{table}[tbp]
\small
\begin{tabular}{lccc}
\toprule
 & \msr & \negc & \wsj  \\ \hline
number of documents            & 1655 & 808   & 582   \\ 
word/doc (mean)           & 148 & 129   & 530   \\ 
sent/doc (mean)      & 7.1 & 5.8   & 25   \\ 
par/doc (mean)      & 2.3 & 2.2   & 10.8   \\ 
referent/doc (mean) & 1 & 2.6   & 15   \\ 
%{|l|}{\textbf{M length sentence}} & 25.8 & 25.8   & 29.5   \\ \hline
number of RE & 11705 & 8378  & 25400  \\ 
%{|l|}{Mean N of mentions / chain}        & 7   & -   \\ \hline
description \%               & 13.84\% & 4\%   & 38.29\%   \\ 
proper name \%                 & 38.09\% & 40.79\%   & 34.57\%   \\ 
pronoun \%              & 41.79\% & 48.75\%   & 27.14\%   \\ 
empty \%              & 6.28\% & 6.47\%   & -   \\ \hline
\end{tabular}
\caption{\label{tab:corpora}
Comparison of the \msr, \negc, and \wsj corpora in terms of their length-related characteristics and distribution of REs. \textit{Doc}, \textit{sent} and \textit{par} stands for \textit{documents}, \textit{sentences} and \textit{paragraphs}.
}
\end{table}

\paragraph{Data Splits.} We have made a document-wise split of the data. We split the \wsj~data in accordance with the CoNLL 2012 Shared Task \citep{pradhan-etal-2012-conll}. Our \wsj training, development, and test sets contain 20275, 2831, and 2294 samples, respectively. We did an 85-5-10 split of the GREC datasets in accordance with \citet{belz2009generating}. After excluding empty pronouns, the \msr~training, development, and test sets contain 9413, 519, 1038 instances, and the \negc~training, development, and test sets contain 6681, 259, 896 instances.

\paragraph{Proportion of Referring Expressions} As shown in Table \ref{tab:corpora}, pronouns and proper names make up 80\% and 89.5\% of the referential instances in \msr and \negc, respectively. This implies that the other two referential forms, namely descriptions and empty references, account for approximately 20\% of the cases in \msr and about 10\% in \negc. Given this imbalance in the frequency of different forms within the two corpora, we question its potential effect on algorithm performance. Specifically, we are wondering if forms with lower frequencies are accurately predicted by the algorithms.

%% file: sec/evaluation.tex
\section{Evaluation} \label{sec:evaluation}

\input{tab/performance.tex}

In this section, we introduce the evaluation protocol and report the performance of the models.

\subsection{Implementation Details} \label{sec:implementation}

For \bert and \roberta, we used \textit{bert-base-cased} and \textit{roberta-base}, both from Hugging Face. For fine-tuning, we set the batch size to 16, the learning rate to 1e-3, the dropout rate to 0.5, and the size of the output layer to 256. We ran each model for 20 epochs and used the one that achieved the highest F1 score on the development set. The implementation details of the classic ML-based models can be found in Appendix~\ref{sec:appendixML}.

\subsection{Evaluation Protocol} \label{sec:protocol}

The main evaluation metric in the GREC-MSR shared tasks was accuracy. 
In addition to accuracy, we also report macro-F1 and weighted-macro F1. We argue that different metrics evaluate algorithms from different perspectives and provide us with different meaningful insights. 
For pragmatic tasks like REG, it makes sense to ask how well an algorithm performs on naturally distributed data which is often imbalanced. For these cases, reporting accuracy and weighted F1 are logical. 
Furthermore, analogous to other classification tasks, minority categories should not be overlooked. Take as an example the class \emph{description} in the \negc corpus, which occurs only 4\%. If a model fails to produce this class, the produced document might sound unnatural. Therefore, it is important to ensure that an algorithm is not over- or under-generating certain classes. Looking into accuracy and macro-F1 together provides insights into such cases.

\subsection{Performance of the Models}\label{subsec:overallacc}

The overall accuracy of the models, their macro F1, and their weighted-macro F1 are presented in Table \ref{tab:performance}. 
We also present the ranking of the models based on these scores in Appendix~\ref{sec:app_rank}.

\paragraph{PLM-based Models.} The best-performing models across all corpora and metrics are PLM-based models.  In six out of nine rankings, \bert and \roberta are ranked as the top two models. The sole exception is \negc, where \bert is the second worst model. The benefit of using PLMs is the largest on the \wsj corpus. For example, \roberta improves the macro F1 score from 69.63 (i.e., the performance of the best ML-based model) to 82.70.

\paragraph{ML-based Models.} In contrast to the robust performance of the PLM models, the performance of the classic ML models is more corpus-dependent. In the case of \msr and \negc, \icsi is the best-performing model, while in the case of \wsj, it is at the bottom section of the rankings. Another interesting observation is the performance of the \udel models. In terms of accuracy, \udel has the highest performance in \negc, while it has the lowest performance in both \msr and \wsj. In terms of macro-F1 rankings, the \negc \udel model dropped from first to last place, whereas \bert improved from penultimate place to second place. In general, our ML models yielded lower scores than the original models used in the GREC study \citep{belz2009generating}. This could be attributed to a variety of factors, including differences in feature engineering and model parameters.

\paragraph{Comparing Different Metrics.} 

Upon comparing average scores across the three metrics, we observe that for \msr and \negc, PLMs are clear winners only when macro-F1 is the metric in question. However, for \wsj, PLMs are winners on all three metrics. This may be because the distribution of categories in \wsj is much more balanced than in the other two corpora.

%% file: tab/performance.tex
\begin{table*}[tbp]
\centering
\small
\begin{tabular}{cccccccccc}
\toprule
& \multicolumn{3}{c}{\msr} & \multicolumn{3}{c}{\negc} & \multicolumn{3}{c}{\wsj} \\
& Acc. & F1 & wF1 & Acc. & F1 & wF1 & Acc. & F1 & wF1 \\ \cmidrule(lr){2-4} \cmidrule(lr){5-7} \cmidrule(lr){8-10} 
\udel & 66.86 & 56.76 & 64.3 & \textbf{80.80} & 55.45 & 77.9 & 63.74 & 64.23 & 63.2 \\
\icsi & \underline{71.19} & 64.73 & 70.4 & 80.36 & 64.53 & \underline{78.6} & 64.62 & 64.15 & 63.4 \\
\cnts & 68.59 & 61.39 & 67.2 & 78.68 & 61.62 & 76.8 & 64.31 & 64.59 & 64.4 \\
\osu & 68.02 & 60.28 & 66.6 & 79.24 & 57.04 & 76.5 & 69.20 & 69.63 & 68.9 \\
\isg & 67.05 & 58.83 & 65.3 & 77.34 & 59.52 & 75.6 & 69.15 & 69.35 & 69.2 \\ \midrule
\bert & \textbf{71.68} & \underline{66.70} & \textbf{71.4} & 77.79 & \underline{72.87} & 77.7 & \underline{80.95} & \underline{80.93} & \underline{80.9} \\
\roberta & 70.91 & \textbf{67.53} & \underline{70.7} & \textbf{80.80} & \textbf{77.29} & \textbf{80.7} & \textbf{82.61} & \textbf{82.70} & \textbf{82.6} \\ \midrule
Average & 69.19 & 62.32 & 67.99 & 79.29 & 64.05 & 77.69 & 70.65 & 70.80 & 70.37 \\
\bottomrule
\end{tabular}
\caption{\label{tab:performance} Overall accuracy (Acc.), macro-averaged F1 (F1), and weighted-macro F1 (wF1) scores of the algorithms depicted in Section~\ref{sec:algorithm}. For instance, \msr-\udel refers to a C5.0 classifier trained on the \msr~corpus, using the feature set mentioned in \citet{greenbacker-mccoy-2009-udel}.}
%Its Acc., F1 and wF1 of this model are 66.86, 56.76, and 64.3, respectively.}
\end{table*}

%% file: sec/analysis.tex
\section{Analysis} \label{sec:analysis}

% Overview of analyses and their motivation
To further compare the different models and investigate the impact of the choice of corpus, we conduct (1) a Bayes Factor (BF) analysis to determine whether the accuracy rates reported in Section~\ref{sec:evaluation} come from similar or different distributions, (2) a per-class evaluation of predictions to assess the success of each model in predicting individual classes, (3) a correlation analysis to quantify how the evaluation results change with respect to the choice of a corpus, and (4) a feature selection study to check how the importance of each feature changes as a function of the choice of corpus.

\subsection{Bayes Factor Analysis} \label{sec:bf}

\input{tab/perclass.tex}

Given that the accuracy scores are provided for all GREC systems in \citet{belz2009generating}, we chose to focus our analysis on the raw distributions of these scores. Our aim is to determine if there are significant differences between the accuracies of our models by comparing these distributions.
We conduct a Bayes Factor analysis with a beta distribution of 0.01 (henceforth: the threshold). This analysis aims to assess, for each pair of accuracies, how strong the evidence is that they come from a common distribution, or from different ones.
A difference below the threshold indicates that accuracy rates come from similar distributions; whereas, a difference above the threshold indicates that they come from different distributions, thus signalling that they differ evidentially. 
We interpret the strength of the evidence in favour of/against similar/different distributions according to \citet{kass1995bayes}. Therefore, based on this approach, we expect that the raw accuracy distributions of the best- and worst-performing models for each corpus differ evidentially. 

For \msr, the comparison between the best- and worst-performing models, namely \bert and \udel, provides no evidence that their accuracy rates are evidentially different from each other (BF = 1.4). The same holds for \negc, where the comparison of the best (\udel and \roberta) and worst (\isg) models appear to have similar probability distributions; therefore, these models are not evidentially different from each other. Conversely, in the case of \wsj, the BF analysis provides strong evidence that the accuracy distributions of the top-performing models, \bert and \roberta, are different from those of the classic ML models.

To summarise, we only observed significant differences in the \wsj-based models; the GREC models show more or less the same accuracy distributions. A reason might be that the aggregated calculation of accuracy loses the specificity of the classes being calculated.

\subsection{Per-class Evaluation} \label{sec:perclass}

As mentioned earlier, the \negc models demonstrate high accuracy (e.g. the highest average accuracy), but we observe a sharp decline in their macro-F1 values. In this analysis, we want to investigate whether the accuracy scores reported in Table~\ref{tab:performance} truly reflect the success of these algorithms or if they are merely the by-product of over-generating the dominant label or under-generating the less frequent label. Table~\ref{tab:perclass} presents the {\em per-class} precision, recall, and F1 scores of these models.

Upon comparing the F1 scores for the class \textit{description} across the three corpora, we observe that the \wsj models consistently achieve the highest scores, with all algorithms exceeding an F1 score of 50. In contrast, the F1 scores for both \msr and \negc are considerably lower than those of \wsj. The F1 scores for \negc are particularly low, with two notable instances, \udel and \osu, scoring 0 and below 10 respectively.  The poor prediction of the class description by the classic ML \negc models is likely due to an insufficient number of instances in the training dataset, thereby hindering the proper training of the algorithms. In contrast, the two PLM models demonstrate acceptable performance in predicting the class description (\bert = 61.16 \& \roberta = 69.02). This could indicate that pre-trained language models are advantageous where there is a class imbalance.

Another interesting observation concerns the high recall of the ``pronoun" prediction in the \negc models. Four of the classic models have a recall of over 92. In the case of \osu, for example, the recall is 95, which means that of all the cases that are pronouns, 95\% are labelled correctly. This is possibly an indication that pronouns have been over-generated in this system. In the PLM models, the recall is below 84.

In sum, the results of our per-class evaluation show the difficulties that the classic ML-based \negc models had in predicting the class \emph{description}. The \msr models also had poor performance in predicting descriptions, yet they were more successful than \negc. These results tentatively suggest that feature-based classification models need to be trained on an adequate and relatively balanced number of instances to reliably predict all classes. The results of this study suggest that the PLM models are less dependent on the choice of corpus, and therefore predict classes more robustly. 

\subsection{Correlation Analysis}

\input{tab/correlation.tex}

To quantify how the evaluation results change with respect to corpora, we compute the Spearman correlation coefficient between every pair of corpora, indicating how the rank of the models changes. Table~\ref{tab:correlation} shows the computed coefficients along with the p-values of the tests. It is noteworthy that only the results evaluated by the macro-weighted F1 on \msr and \negc are significantly correlated ($p < .001$). 

The lack of correlation between the results on \msr/\wsj and those on \negc/\wsj suggests that using a corpus of a different genre could greatly influence the ranking of the models and, therefore, make the conclusions difficult to generalise.
Additionally, these results are in line with the fact that \msr and \negc are from the same source, both being the introductory part of Wikipedia articles, and a higher correlation is to be expected. Also, we may conclude that macro-averaged F1 is a more reliable evaluation metric (see the discussions in Section~\ref{sec:evaluation}, Section~\ref{sec:bf}, and Section~\ref{sec:perclass}). 

\subsection{Feature Selection Study}

We performed a feature importance analysis to check whether the contribution of linguistic factors changes depending on the choice of the corpus. We used XGBoost from the family of Gradient Boosting trees~\citep{xgboost2016} and then computed the permutated variable importance for each model. Data were analysed in two ways: firstly, we used the complete dataset, as outlined in Section \ref{sec:corpus}; secondly, we excluded first-mention REs to concentrate only on subsequent mentions. Considering that the choice of a referent' first mention is less context-dependent, we only report on the latter dataset below:
\begin{figure*}[t]
    \centering
    \includegraphics[scale=0.07]{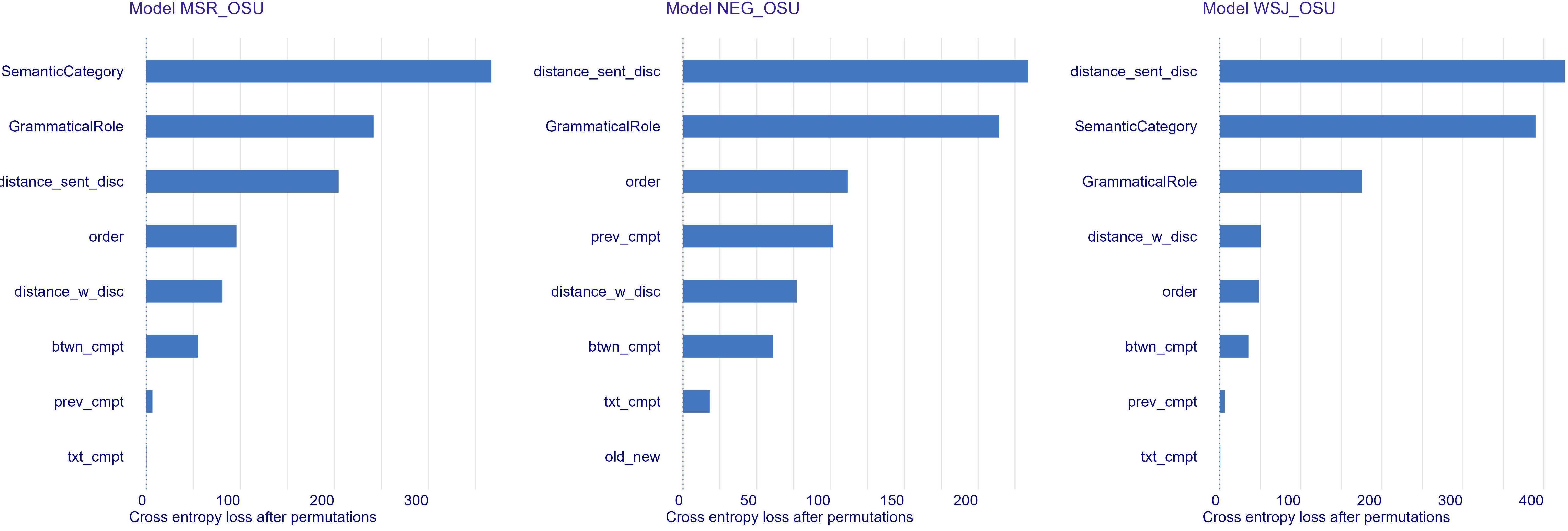}
    \caption{Different rankings of the features in \msr, \negc, and \wsj \osu models.}
    \label{fig:rankings}
\end{figure*}

As expected, the ranking of feature importance varies across different corpora.  However, a substantial overlap is observed when considering the most important features across the three corpora. An example is the semantic category of the REs that is used in various \msr and \wsj REG models.\footnote{Only human referents are annotated in \negc; therefore, this feature is not applicable.} In the case of \msr, the REs belong to five semantic categories: human, city, country, river, and mountain. In the case of \wsj, the REs are annotated for a wide range of categories including human, city, country, organisation, objects, etc.
Notably, in every model that employs semantic category information, this feature has either the highest or second-highest importance ranking. A plausible explanation could be that humans use different referencing strategies to refer to different categories of referents. 

In addition to the semantic category, the grammatical role of the RE and the categorical sentential distance to the antecedent consistently have a high importance ranking. The grammatical role marks the distinction between subject, object, and determiner roles. The categorical distance in the number of sentences provides information on how far an RE is to its nearest coreferential antecedent. For instance, whether they are both in the same sentence or are separated by one or more sentences. Figure \ref{fig:rankings} illustrates the importance rankings of the \osu features in the three corpora. Other importance ranking graphs are available in Appendix \ref{sec:appendixrankings}. For a comprehensive description of all features employed in classic ML models and the feature importance analysis, refer to \citet{same-van-deemter-2020-linguistic}.

%% file: tab/perclass.tex
\begin{table*}[ht]
\small
\centering
\begin{tabular}{llcccccccccc}
  \toprule
  & & \multicolumn{3}{c}{\textsc{msr}} & \multicolumn{3}{c}{\textsc{neg}}  & \multicolumn{3}{c}{\textsc{wsj}} \\ \cmidrule(lr){3-5} \cmidrule(lr){6-8} \cmidrule(lr){9-11}
Model & Category & P & R & F & P & R & F & P & R & F\\ 
  \midrule
\multirow{3}{*}{\texttt{Udel}} & description & 55.36 & 19.38 & 28.71 & 0.00 & 0.00 & 0.00 & 60.29 & 62.95 & 61.59 \\ 
  & name & 72.39 & 62.21 & 66.92 & 76.65 & 80.32 & 78.44 & 60.42 & 49.44 & 54.38 \\ 
  & pronoun & 64.53 & 88.51 & 74.64 & 84.06 & 92.14 & 87.91 & 71.00 & 83.44 & 76.72 \\ 
  \midrule
\multirow{3}{*}{\texttt{ICSI}} & description & 51.69 & 38.12 & 43.88 & 100.00 & 17.74 & 30.13 & 81.92 & 40.53 & 54.22 \\ 
  & name & 80.33 & 66.82 & 72.95 & 81.85 & 73.14 & 77.25 & 55.12 & 86.40 & 67.37 \\ 
  & pronoun & 69.41 & 87.39 & 77.37 & 79.05 & 94.76 & 86.19 & 72.17 & 69.61 & 70.86 \\ 
  \midrule
\multirow{3}{*}{\texttt{CNTS}} & description & 53.68 & 31.88 & 40.00 & 75.00 & 14.52 & 24.33 & 64.31 & 63.67 & 63.30 \\ 
  & name & 76.79 & 61.75 & 68.45 & 77.84 & 72.87 & 75.27 & 60.34 & 66.75 & 63.38 \\ 
  & pronoun & 66.16 & 88.51 & 75.72 & 79.32 & 92.14 & 85.25 & 71.90 & 62.54 & 66.89 \\ 
\midrule
\multirow{3}{*}{\texttt{OSU}} & description & 53.57 & 28.12 & 36.88 & 100.00 & 4.84 & 9.23 & 72.70 & 56.91 & 63.84 \\ 
& name & 69.39 & 68.43 & 68.91 & 79.01 & 72.07 & 75.38 & 63.56 & 73.30 & 68.08 \\ 
& pronoun & 69.20 & 81.98 & 75.05 & 79.27 & 95.20 & 86.51 & 73.43 & 80.87 & 76.97 \\ 
\midrule
\multirow{3}{*}{\texttt{ISG}} & description & 57.97 & 25.00 & 34.93 & 77.78 & 11.29 & 19.72 & 73.88 & 63.41 & 68.25 \\ 
& name & 71.46 & 65.21 & 68.19 & 71.77 & 79.79 & 75.57 & 62.19 & 76.64 & 68.66 \\ 
& pronoun & 65.10 & 84.01 & 73.36 & 82.30 & 84.28 & 83.28 & 75.36 & 67.36 & 71.14 \\ 
\midrule
\multirow{3}{*}{\texttt{BERT}} & description & 52.86 & 46.25 & 49.33 & 62.71 & 59.68 & 61.16 & 82.63 & 79.37 & 80.97 \\
& name & 74.35 & 72.81 & 73.57 & 77.32 & 75.27 & 76.28 & 79.64 & 82.69 & 81.14 \\
& pronoun & 74.84 & 79.73 & 77.21 & 80.04 & 82.31 & 81.16 & 80.48 & 80.87 & 80.67 \\ 
\midrule
\multirow{3}{*}{\texttt{RoBERTa}} & description & 56.33 & 55.62 & 55.97 & 76.47 & 62.90 & 69.02 & 86.19 & 77.40 & 81.56 \\
& name & 76.50 & 64.52 & 70.00 & 78.70 & 80.59 & 79.63 & 77.22 & 89.25 & 82.80 \\
& pronoun & 71.40 & 82.66 & 76.62 & 83.04 & 83.41 & 83.22 & 86.47 & 81.19 & 83.75 \\ 
   \bottomrule
\end{tabular}
\caption{\label{tab:perclass} Per-class precision, recall and F1 score of each label. The results report on training seven different algorithms on three corpora for predicting three labels, namely description, name, and pronoun.}
\end{table*}

%% file: tab/correlation.tex
\begin{table}[t]
\centering
\begin{tabular}{llccc}
\toprule
 & & acc & F1 & wF1\\
\midrule
\multirow{2}{*}{\makecell[c]{\msr/\negc}} & $r_s$ & -0.1081 & 0.9643 & 0.4643 \\
& $p$ & 0.8175 & 0.0005 & 0.2939 \\ \midrule
\multirow{2}{*}{\makecell[c]{\msr/\wsj}} & $r_s$ & 0.2857 & 0.5357 & 0.4643 \\
& $p$ & 0.5345 & 0.2152 & 0.2939 \\ \midrule
\multirow{2}{*}{\makecell[c]{\negc/\wsj}} & $r_s$ & -0.1261 & 0.5000 & -0.0357 \\
& $p$ & 0.7876 & 0.2532 & 0.9394 \\
\bottomrule
\end{tabular}
\caption{Spearman correlation coefficient $r_s$ and the p-value between every pair of corpora in terms of accuracy, macro-averaged F1, and weighted F1.}
\label{tab:correlation}
\end{table}

%% file: sec/discussion.tex
\section{Discussion} \label{sec:discussion}

In this paper, we have conducted a series of reproductions, evaluations, and analyses to check whether the conclusions of GREC are still true after 15 years. Below, we summarise and discuss our findings in accordance with our three research questions in Section~\ref{sec:rq}. We also report our post-hoc observations on the choice of evaluation metric.

\paragraph{Performance of REG Algorithms.}

To answer research question $\mathcal{R}_2$, we extended the GREC by introducing a corpus of a different genre, \wsj, and two pre-trained (PLM-based) REG models. We found that, on \msr, PLM-based and ML-based models perform similarly, as confirmed by both the BF and per-class analyses.
With regards to \negc, PLM-based and ML-based models have similar accuracy scores, as confirmed by the BF analysis, but there are large differences when micro-F1 is used, as confirmed by the per-class evaluation (i.e., ML-based models have difficulty predicting descriptions). On \wsj, PLM-based models are the clear winners.

These results suggest that, in terms of explanatory power, PLM-based models have good performance and good ``direct support'', i.e., a good ability to generalise to different contexts (see \citet{van2022role} for further discussion). Whether they have good ``indirect support'' (e.g., whether their predictions are in line with linguistic theories) needs to be investigated in further probing studies. 
 
\paragraph{Impact of the Choice of Corpus.}

As our evaluations and analyses demonstrate, the choice of corpus plays a crucial role in assessing REG algorithms. This role is twofold. Firstly, the choice of corpus strongly influences the evaluation results, pertaining to the research question $\mathcal{R}_1$. Secondly, in addition to the score differences discussed in Section~\ref{sec:evaluation}, we found that: (1) the difference between PLM-based and ML-based models on \wsj is larger (and evidentially different) than on \msr and \negc models (as evidenced by the BF analysis); (2) the correlations of the evaluation results between \wsj and both \msr and \negc are not significant.

For $\mathcal{R}_3$, we conducted feature selection analyses across the three corpora, discovering that the importance of the features ranks differently for each corpus. This suggests that when investigating the ``indirect support'' for a model, one needs to aggregate findings from multiple corpora with different genres.

\paragraph{The Use of Evaluation Metrics.}

As we discussed in Section~\ref{sec:protocol}, different metrics evaluate different aspects of a model. This was further ascertained by the inconsistency of the BF analysis and per-class analysis. 
One lesson we have learned is that it is not enough to report or do analyses on a single metric. Another lesson is that the evaluation results by macro-F1 are more reliable than other metrics because (1) they are consistent across corpora with similar genres (i.e., \msr and \negc; see the Correlation analysis results); (2) the differences identified by using macro-F1 can be confirmed by the per-class evaluation.

\section{Conclusion}

We are now in a position to address the question that we raised in the Introduction: Can the conclusions from the GREC shared tasks still be trusted? By examining a wider class of corpora, models, and evaluation metrics than before, we found that the answer to this question is essentially negative since the GREC conclusions are prone to drastic change once a different corpus or a different metric is employed.

Perhaps this should come as no surprise. According to a widely accepted view of scientific progress (e.g., \citet{jayn}; applied to NLP in \citep{van2022role}), theories should be updated again and again in light of new data (i.e., indirect Support), and when new models are proposed, the plausibility of existing models should be compared against the plausibility of these new models (as well as pre-existing ones). New metrics deserve a place in this story as well, even though they are often overlooked. 
In other words, what we have seen in the present study is nothing more than science in progress -- something we are bound to see more of as the enterprise called NLP-as-Science matures.

%% file: sec/appendix.tex
\section{Ranking of the Models} \label{sec:app_rank}

\paragraph{Accuracy-based Ranking}
\mbox{}\\

\noindent \msr : \bert $>$ \icsi $>$ \roberta $>$ \cnts $>$ \osu $>$ \isg $>$ \udel 

\noindent\negc: \udel $=$ \roberta $>$ \icsi $>$ \osu $>$ \cnts $>$ \bert $>$ \isg

\noindent \wsj: \roberta $>$ \bert $>$ \osu $>$ \isg $>$ \icsi $>$ \cnts $>$ \udel

\paragraph{Macro-F1 Ranking}
\mbox{}\\

\noindent \msr : \roberta $>$ \bert  $>$ \icsi $>$ \cnts $>$ \osu $>$ \isg $>$ \udel

\noindent\negc: \roberta $>$ \bert $>$ \icsi $>$ \cnts $>$ \isg $>$ \osu $>$ \udel

\noindent \wsj: \roberta $>$ \bert $>$ \osu $>$ \isg $>$ \cnts $>$ \udel $>$ \icsi

\paragraph{Macro-weighted F1 Ranking}
\mbox{}\\

\noindent \msr : \bert  $>$ \roberta $>$ \icsi $>$ \cnts $>$ \osu $>$ \isg $>$ \udel 

\noindent\negc: \roberta $>$ \icsi $>$ \udel $>$ \bert $>$ \cnts $>$ \osu $>$ \isg

\noindent \wsj: \roberta $>$ \bert $>$ \isg $>$ \osu $>$ \cnts  $>$ \icsi $>$ \udel
\mbox{}\\

\section{Implementation Details for ML-based Models} \label{sec:appendixML}

The R programming language was used mostly for running the classic ML  models. The specification of the models can be found below: 

\paragraph{Conditional Random Field [CRF].} The R Package CRF (\url{https://cran.r-project.org/web/packages/crfsuite/}) was used to train these models. The iterations are set to 3000, and the learning method is Stochastic Gradient Descent with L2 regularization term (l2sgd).

\paragraph{Decision Tree [C5.0].} The R Package C5.0 \citep{kuhn2018package} was used to build the decision trees. The number of boosting iterations (trials) is set to 3, and the splitting criterion is information gain (entropy).

\paragraph{Memory-Based Learning [MBL].} As mentioned before, we implemented the k-Nearest Neighbors [KNN] algorithm instead of MBL. The R package caret with the method KNN was used to implement this model.

\paragraph{Maximum Entropy [MaxEnt].} The multinom algorithm from the nnet R package was used to implement this model. 

\paragraph{Multi-Layer Perceptron [MLP].} The Keras package was used to implement MLP. The model consists of two hidden layers with 16 and 8 units, respectively. The hidden layers use the rectified linear activation function (ReLU), and the output layer uses the Sigmoid activation function. The model is fitted for 50 training epochs. In addition, 50 samples (batch size) are propagated through the network.

\paragraph{eXtreme Gradient Boosting [XGBoost].} XGBoost was used for the feature selection experiments. We used the R packages xgboost and DALEXtra for the analysis. We set the learning rate to 0.05, the minimum split loss to 0.01, the maximum depth of a tree to 5, and the sub-sample
ratio of the training instances to 0.5.

\section{Feature Importance Rankings} \label{sec:appendixrankings}
\begin{figure*}[t]
    \centering
    \includegraphics[scale=0.068]{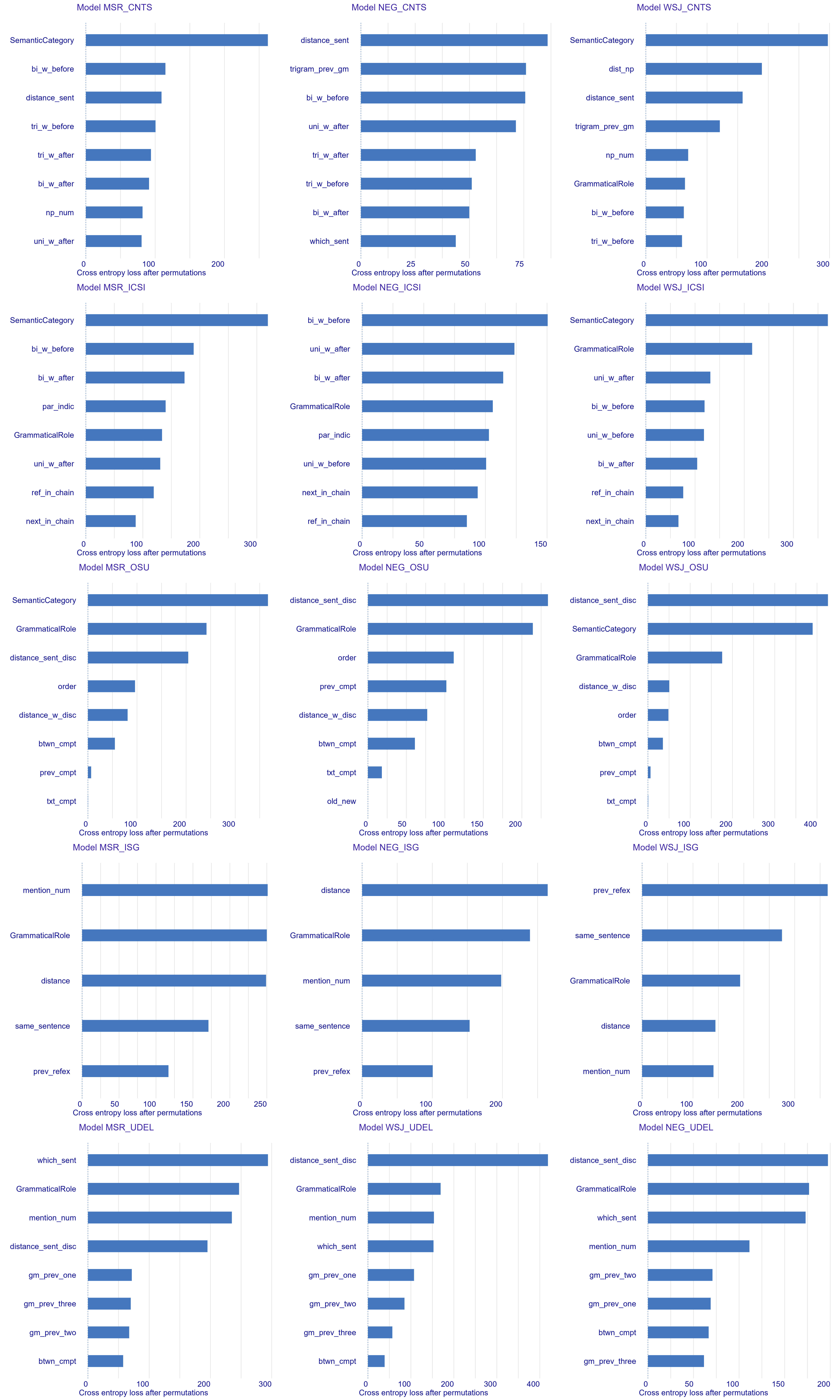}
    \caption{Importance ranking of the features in \msr, \negc, and \wsj models.}
    \label{fig:allrankings}
\end{figure*}

The graphs in Figure \ref{fig:allrankings} show the rankings across \msr, \wsj, and \wsj. A maximum number of eight features is depicted in the graphs.
